\documentclass[times, review, 10pt]{elsarticle}
\usepackage{caption}

\usepackage{amssymb}
\usepackage{amsmath}
\usepackage{verbatim}
\usepackage{graphicx}
\usepackage{color} 
\usepackage{bbding} 
\usepackage{setspace}
\definecolor{markcolor}{RGB}{0, 0, 0} 
\definecolor{agreen}{rgb}{0.0, 0.5, 0.0}
\definecolor{ashgrey}{rgb}{0.7, 0.75, 0.71}
\usepackage{balance}
\usepackage{multirow}
\usepackage{booktabs}
\usepackage{hyperref}
\usepackage{subcaption}
\usepackage{makecell}

\journal{Pattern Recognition}

\begin{document}

\begin{frontmatter}



\title{Advancing Metallic Surface Defect Detection via Anomaly-Guided Pretraining on a Large Industrial Dataset}

\author[ad1]{Chuni Liu}
\author[ad1]{Hongjie Li}
\author[ad1]{Jiaqi Du}
\author[ad1]{Yangyang Hou}
\author[ad1]{Qian Sun}
\author[ad1]{Lei Jin}
\author[ad1]{Ke Xu \corref{BCA}}

\address[ad1]{Collaborative Innovation Center of Steel Technology, University of Science and Technology Beijing, China.}

\cortext[BCA]{corresponding authors: xuke@ustb.edu.cn.}


\begin{abstract}
The pretraining-finetuning paradigm is a crucial strategy in metallic surface defect detection for mitigating the challenges posed by data scarcity. However, its implementation presents a critical dilemma. Pretraining on natural image datasets such as ImageNet, faces a significant domain gap. Meanwhile, naive self-supervised pretraining on in-domain industrial data is often ineffective due to the inability of existing learning objectives to distinguish subtle defect patterns from complex background noise and textures. 
To resolve this, we introduce Anomaly-Guided Self-Supervised Pretraining (AGSSP), a novel paradigm that explicitly guides representation learning through anomaly priors. AGSSP employs a two-stage framework: (1) it first pretrains the model's backbone by distilling knowledge from anomaly maps, encouraging the network to capture defect-salient features; (2) it then pretrains the detector using pseudo-defect boxes derived from these maps, aligning it with localization tasks. To enable this, we develop a knowledge-enhanced method to generate high-quality anomaly maps and collect a large-scale industrial dataset of 120,000 images. Additionally, we present two small-scale, pixel-level labeled metallic surface defect datasets for validation. Extensive experiments demonstrate that AGSSP consistently enhances performance across various settings, achieving up to a 10\% improvement in mAP@0.5 and 11.4\% in mAP@0.5:0.95 compared to ImageNet-based models. All code, pretrained models, and datasets are publicly available at \href{https://clovermini.github.io/AGSSP-Dev/}{https://clovermini.github.io/AGSSP-Dev/}.

\end{abstract}


\end{frontmatter}



\section{Introduction}
\label{sec:introduction}

Metallic surface defect detection is critical for industrial quality control~\cite{lv2020deep}, as even a single missed defect may lead to substantial economic losses. While manual inspection has been widely adopted, it is time-consuming and error-prone, motivating the use of deep learning-based methods for automated defect detection~\cite{liu2025global}. However, their effectiveness is severely constrained by the scarcity and high variability of defect samples in real-world industrial scenarios. For example, in steel pipe production, defects account for less than 0.1\% of daily images, making data collection costly and exhibiting an inherently severe class imbalance. Such data limitations hinder the learning of robust defect representations and impair generalization performance~\cite{ZHANG2021107571}. 

\begin{figure}[!t]
\centerline{\includegraphics[width=\columnwidth]{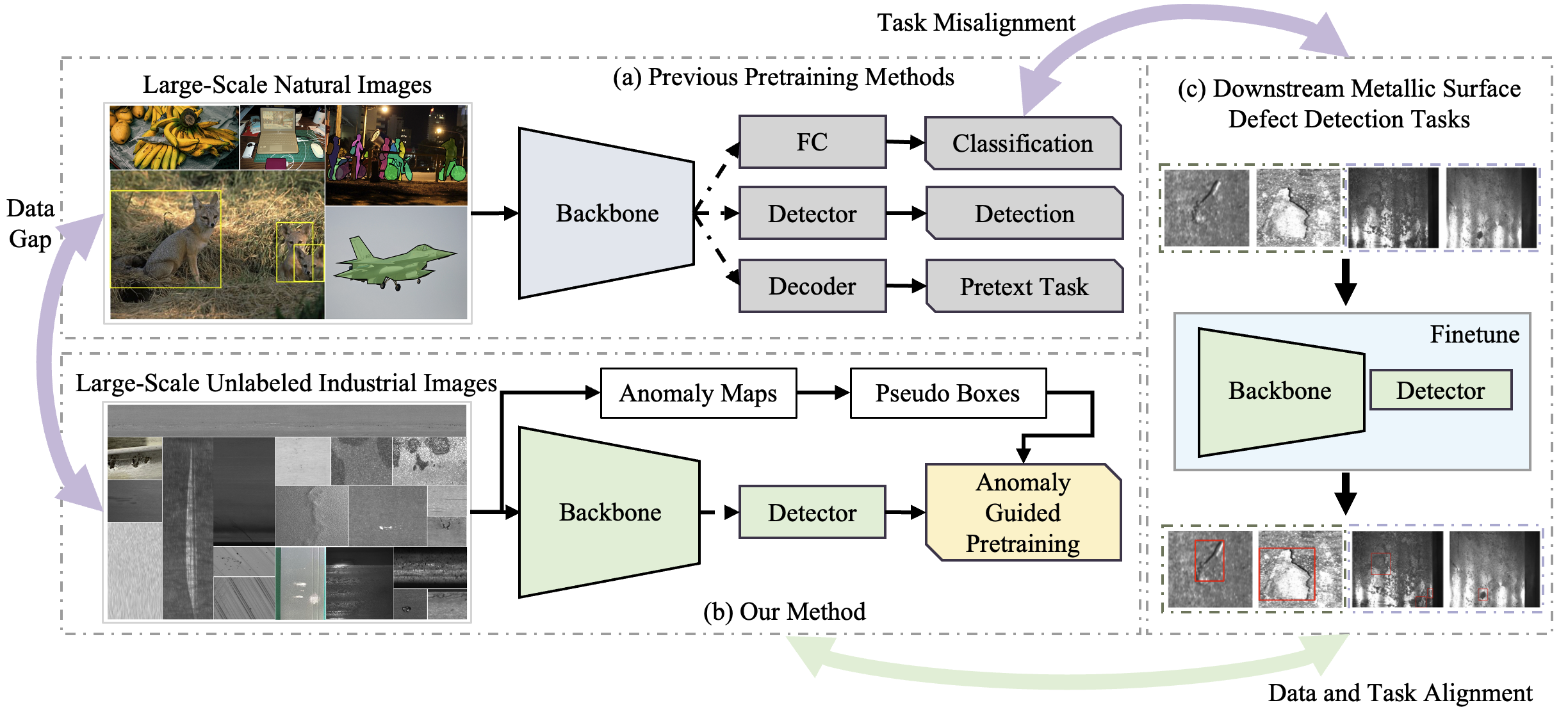}}
\caption{Motivation of our method. 
(a) Most previous pretrained models rely on large-scale natural image datasets with either supervised (e.g., classification, object detection) or self-supervised pretext tasks. These models often experience performance declines on metallic surface defect detection due to data distribution differences and misaligned training objectives.
(b) Our method is pretrained on a large-scale unlabeled industrial dataset, utilizing anomaly maps to guide the pretraining of both the model backbone and detector for downstream tasks.
(c) The downstream tasks are finetuned based on the pretrained model.}
\label{fig:abstract}
\end{figure}

Researchers have explored various strategies to improve deep learning performance under limited data conditions. One line of work focuses on generating synthetic defect samples using techniques such as CutPaste~\cite{li2021cutpaste} or diffusion-based models~\cite{song2025defectfill,s25103038}. However, faithfully reproducing the complexity and variability of real-world defects remains challenging, limiting the effectiveness of these approaches. In parallel, transfer learning, particularly the pretraining–finetuning paradigm, has long served as an established solution. Models pretrained on large-scale datasets, including ImageNet~\cite{deng2009imagenet} and COCO~\cite{lin2014microsoft}, as well as recent self-supervised learning (SSL) models~\cite{chen2021empirical,xie2022simmim,oquab2024dinov2}, have demonstrated strong generalization across a wide range of downstream tasks. Nevertheless, their effectiveness in metallic surface defect detection is limited by the domain gap between natural and industrial images, along with the misalignment between pretraining objectives and defect detection requirements, as illustrated in Figure~\ref{fig:abstract}. 

Large-scale supervised pretraining on industrial defect datasets can theoretically provide task-aligned features, but obtaining the necessary bounding-box annotations is impractical due to the scarcity and imbalance of defect samples. A more feasible alternative is SSL on large-scale, unlabeled, domain-specific industrial datasets, which bypasses costly annotations while capturing domain-relevant representations for defect detection. To this end, we constructed an unlabeled dataset comprising 120,000 industrial surface images spanning 61 categories for training and annotated two small-scale metallic surface defect datasets for validation. Counterintuitively, our experiments reveal that standard in-domain SSL methods often underperform ImageNet-pretrained baselines. We attribute this limitation to the representation-learning objectives underlying current SSL methods~\cite{HU2023110578, torpey2024large, liu2023tower}, which focus on contrastive learning or reconstruction. While effective for general-purpose features, these objectives struggle to capture subtle, low-contrast defects that are easily confounded with noise, complex textures, and variable lighting on metallic surfaces. Evidently, simply enlarging in-domain datasets is insufficient. The key challenge lies in guiding pretraining to focus on anomalous regions that characterize defects. Building on this insight, we propose leveraging anomaly maps, pixel-wise heatmaps that indicate potential defect regions, as prior knowledge to explicitly guide representation learning. 

Besides, most SSL studies focus on pretraining the model backbone, leaving the detector head (i.e., the neck and head) randomly initialized. This oversight creates a misalignment with the downstream detection task and can significantly degrade performance, particularly in data-scarce scenarios. Recent studies in object detection~\cite{li2023aligndet, huang2024mutdet} have demonstrated that pretraining the detector yields substantial performance gains over pretraining the backbone alone. 
Therefore, we introduce a two-stage anomaly-guided self-supervised pretraining framework, which jointly addresses both backbone feature representation and detector task alignment using a unified, anomaly-guided approach. It comprises two consecutive stages: Anomaly Map Guided Backbone Pretraining (AGBP) and Anomaly Box Guided Detector Pretraining (AGDP). In the first stage, knowledge from anomaly maps is distilled into the backbone’s high-level features, directing the model’s attention toward potential defects. This distillation process can be seamlessly integrated with other pretraining tasks to enhance representation quality. In the second stage, pseudo-defect boxes are derived directly from these same anomaly maps to pretrain the detector. This enables the model to learn the concept of anomalous objects and efficiently initialize all components of the detection model using a unified, anomaly-focused signal.

To generate the required anomaly maps, we propose the unsupervised Knowledge-enhanced Anomaly Detection (KeAD) method, based on the few-shot, multi-modal WinCLIP~\cite{jeong2023winclip} model. Unlike conventional unsupervised anomaly detection approaches, which require separate training on multiple normal images for each scenario, KeAD is well-suited for our large-scale unlabeled dataset, which contains no labels distinguishing normal from defective samples. This method adapts efficiently to new categories with only a few normal samples and allows the incorporation of domain knowledge via text prompts. We introduce two key enhancements over the base WinCLIP model: using detailed defect-specific descriptions instead of generic templates, and integrating Clip Surgery~\cite{li2023clip} to strengthen local feature representation. Importantly, although the generated anomaly maps are probabilistic and may contain noise, we show that this targeted guidance is sufficient for pretraining, enabling the model to learn representations that generalize effectively to downstream defect detection tasks.

Our main contributions are summarized as follows:
\begin{itemize}
    \item We propose a novel two-stage pretraining framework, AGSSP, that leverages anomaly maps to holistically align the representation learning of both the backbone and detector with the downstream task. This approach significantly enhances finetuning performance, particularly in data-scarce settings.
    
    \item We introduce KeAD, an anomaly detection method that supports AGSSP by generating anomaly maps for guided pretraining. By incorporating domain-specific knowledge, KeAD achieves strong few-shot performance, allowing for rapid adaptation across diverse industrial scenarios.

    \item We gather a large industrial dataset containing 120,000 images spanning 61 object categories for pretraining purposes and annotate two small-scale datasets of metallic surface defects for validation, which can facilitate future research.

    \item Comprehensive experiments demonstrate the effectiveness of AGSSP, which supports multiple backbones, detectors, and pretraining methods. It achieves significant improvements across three metallic surface defect datasets and one non-metallic surface dataset, surpassing models pretrained on ImageNet and COCO.
\end{itemize}

The remainder of this article is structured as follows: Section~\ref{sec:related_works} provides a brief overview of related work. Section~\ref{sec:datasets} outlines the datasets we collected. Section~\ref{sec:method} details the proposed methodology. Section~\ref{sec:experiments} presents the experimental findings and our analysis. Section~\ref{sec:conclusion} offers the conclusion.

\section{Related work}
\label{sec:related_works}

\subsection{Metallic Surface Defect Detection}

Advances in deep learning have driven significant progress in object detection algorithms, evolving from anchor-based two-stage frameworks (e.g., Faster R-CNN~\cite{ren2016faster}) to efficient single-stage detectors (e.g., the YOLO family, including the latest YOLOv11~\cite{yolo11_ultralytics} and YOLOv12~\cite{tian2025yolov12}), and more recently to Transformer-based query-driven architectures (e.g., DINO~\cite{zhang2023dino}, RT-DETR~\cite{lv2023detrs}). These advances have demonstrated strong effectiveness in surface defect detection, while tailored variants~\cite{zhou2025mpa} such as SLF-YOLO~\cite{liu2025metal} have further improved efficiency and representation for the specific challenges posed by metallic surfaces. 
Parallel to architectural innovations, considerable research has focused on alleviating the challenge of data scarcity through Few-Shot Learning (FSL). Recent approaches have explored sophisticated architectures that leverage techniques such as multiscale adaptive prototypes to accommodate diverse defect patterns~\cite{huang2025multi}, and semantic priors to enhance the detection of tiny or weak defects~\cite{ma2024spdp}. Despite their progress, these methods primarily optimize network design for finetuning, while continuing to rely on general-purpose pretrained weights. Our method instead focuses on tailored pretraining for defect detection to obtain domain-relevant representations from the outset.

\subsection{Self-Supervised Learning}

SSL offers a powerful paradigm for leveraging unlabeled datasets, which is valuable for defect detection where labeled data are scarce and costly to obtain. Existing SSL methods can be broadly categorized into several paradigms. Contrastive learning approaches, such as MoCov3~\cite{chen2021empirical}, learn discriminative representations by encouraging similarity between positive pairs while enforcing separation from negative samples in the embedding space. Image reconstruction methods, including SimMIM~\cite{xie2022simmim}, focus on restoring masked or corrupted image regions, thereby compelling the model to capture semantic and structural information. Self-distillation techniques, exemplified by DINO~\cite{caron2021emerging,oquab2024dinov2}, adopt teacher–student frameworks with momentum updates to promote stable representation learning without the need for explicit negative samples. 

An increasing number of studies have explored SSL on large-scale, domain-specific unlabeled datasets, motivated by the expectation that in-domain pretraining can outperform conventional ImageNet-based initialization. For example, contrastive learning on 700k radiographs~\cite{zhou2020comparing} and category classification pretraining on over 100,000 microscopy images~\cite{stuckner2022microstructure} have both yielded superior downstream performance. However, such benefits are not universal: a large-scale evaluation~\cite{torpey2024large} found that pretraining on 22,000 electroluminescent images was less effective than ImageNet pretraining for solar cell defect segmentation. It suggests that domain-specific SSL is advantageous only when the learned representations transfer effectively to the target task. As a result, several specialized SSL strategies have been proposed to tackle specific industrial vision tasks. For instance, SCRL-EMD~\cite{HU2023110578} proposed a SimSiam-based framework that enhances contrastive learning for steel defects by addressing noise introduced by random augmentations of multi-target images and is pretrained on 11,681 unlabeled samples. Another method, Tower Masking MIM~\cite{liu2023tower}, employed a masking scheme tailored to the structural characteristics of power lines, leveraging 39,514 unlabeled images. Similarly, a weakly supervised approach~\cite{11028102} used contrastive learning on 25,020 coarsely labeled images to improve the distinction between targets and backgrounds in power line inspections. While these strategies have proven successful, they are highly tailored to specific industrial objects or scenarios. In contrast, our proposed AGSSP framework is built on the more general principle of anomaly guidance, making it broadly applicable across diverse defect detection tasks. Furthermore, our AGSSP framework can be integrated with existing specialized architectures, such as those mentioned above.

Besides, a common limitation of these strategies is their primary emphasis on pretraining the model's backbone, overlooking the detector component in detection models. To overcome this limitation, some researchers have suggested incorporating specialized pretraining tasks for the detector component.~\cite{li2023aligndet} implemented selective search to generate pseudo-labels for natural images, followed by pretraining the detector using a contrastive loss approach. 
~\cite{huang2024mutdet} employed SAM~\cite{kirillov2023segment} to produce object bounding boxes and presented a mutually optimizing pretraining strategy for remote sensing object detection. To the best of our knowledge, we are the first to propose utilizing anomaly map as a form of explicit supervision to guide the SSL process for both the backbone and the detector, specifically tailored to the characteristics of industrial defect data. Moreover, our method is developed on a dataset of 120,000 industrial images, which represents, to our knowledge, the largest pretraining corpus in this domain.

\subsection{Industrial Anomaly Detection}

Conventional anomaly detection methods~\cite{roth2022towards,sun2025png,yu2025attention} primarily relied on modeling normal data distributions to identify deviations, typically necessitating distinct models and training procedures for each application scenario. Recent advancements have further pushed the boundaries of this paradigm: MambaAD~\cite{he2024mambaad} introduces Mamba for multi-class unsupervised anomaly detection; Omni-AD~\cite{quan2025omni} and INP-Former~\cite{luo2025exploring} refine reconstruction efficacy by capturing global-local features and intrinsic normal prototypes, respectively; while DAF~\cite{cai2024discrepancy} enhances the robustness of synthetic data-based detection. Recent benchmark efforts have also begun to consider more realistic and diverse industrial settings~\cite{zhu2026real}. Despite these innovations, the field has recently been transformed by the advent of vision-language models, particularly WinCLIP~\cite{jeong2023winclip}, which leveraged image-text alignment for zero-/few-shot anomaly detection. Building on CLIP's foundation~\cite{radford2021learning}, this paradigm has spawned numerous anomaly detection methods that demonstrate remarkable generalization capabilities across diverse contexts without scenario-specific training. Recent extensions of CLIP for industrial anomaly detection employ auxiliary dataset training with additional trainable parameters: April-GAN~\cite{chen2023zero} (vision linear layer), AnomalyCLIP~\cite{zhou2023anomalyclip} (learnable text prompts), \textcolor{markcolor}{AdaCLIP~\cite{cao2024adaclip} (hybrid prompts)}, and AA-CLIP~\cite{ma2025aa} (dual-encoder adapters with contrastive learning). While these approaches improve performance, they may risk overfitting, whereas our training-free WinCLIP adaptation achieves gains without this issue.

\section{Datasets}
\label{sec:datasets}

\subsection{Large-Scale Industrial Dataset}

In this work, we collect a large-scale industrial dataset comprising 120,000 images for pretraining. Half of the dataset is aggregated from 20 publicly available industrial surface defect datasets, encompassing notable benchmarks such as Severstal~\cite{severstal2019}, FSSD-12~\cite{feng2023cross} and CR7-DET~\cite{chen2024efficient}. The other half originates from real-world data collected from 14 steel plants and production lines, encompassing a wide spectrum of metallic materials, including aluminum plates, steel sheets, steel strips, steel pipes, and steel rails. The dataset includes 61 categories of objects categorized by material and data source, offering a comprehensive representation of diverse surface conditions and textures. Remarkably, the metallic surface datasets used for validation do not overlap with the pretraining dataset, ensuring fair and unbiased evaluation (please see Appendix for detailed data sources).

\begin{figure}[!t]
\centering
\includegraphics[width=0.95 \linewidth]{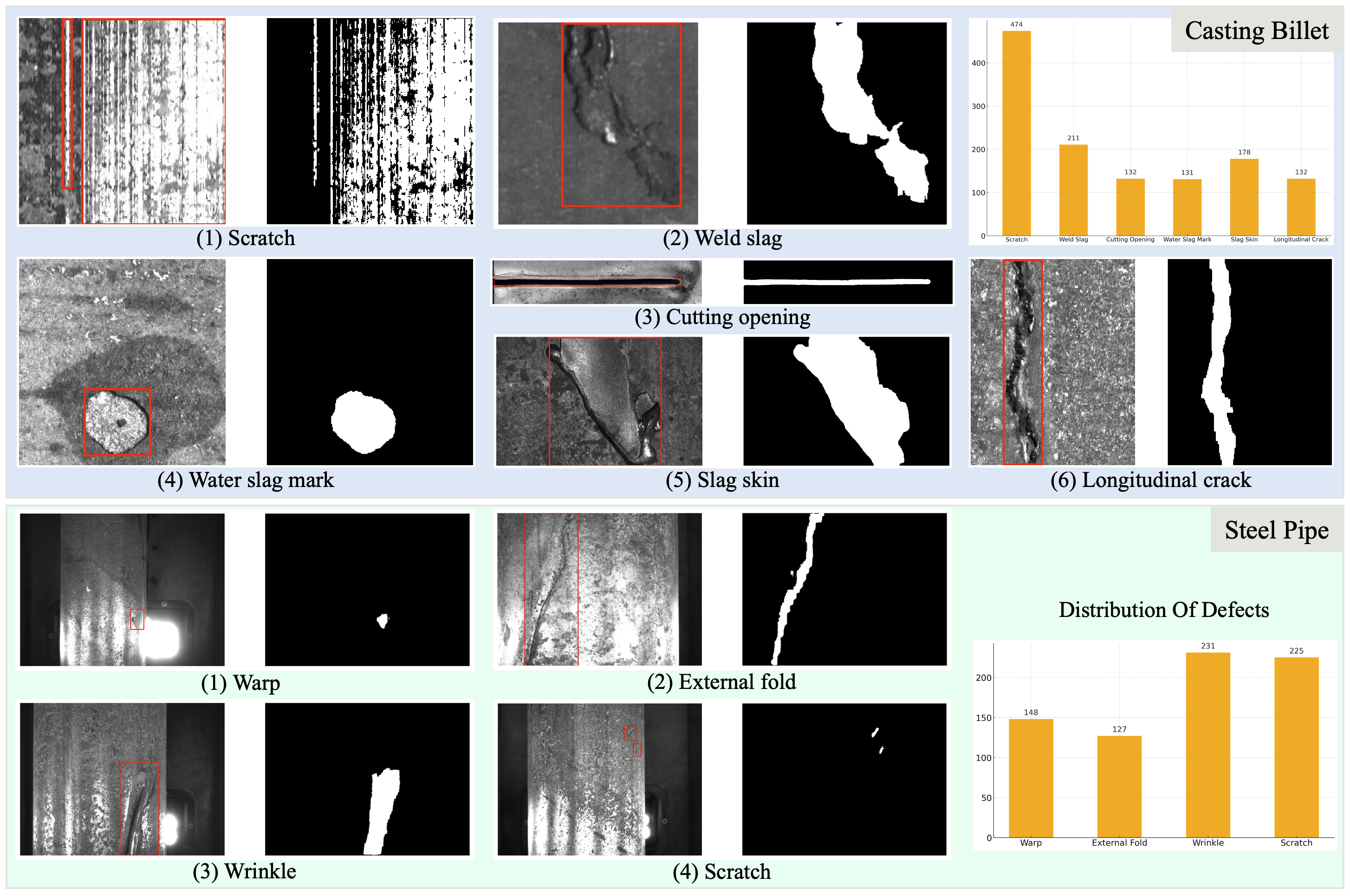}
\caption{Examples of defect images with annotations in Casting Billet and Steel Pipe, along with corresponding defect quantity distributions.}
\label{fig:datasets}
\end{figure}

\subsection{Metallic Surface Defect Datasets}
We curate two small-scale metallic surface defect datasets to assess the downstream performance of pretrained models. For both datasets, pixel-wise annotations are generated with SAM~\cite{kirillov2023segment} using bounding boxes as prompts and subsequently refined through manual correction. Representative samples, corresponding labels, and defect quantity distributions are shown in Figure~\ref{fig:datasets}. 
\subsubsection{Casting Billet} This dataset comprises 1,060 images with resolutions ranging from 96$\times$106 to 3,228$\times$492, depicting high-temperature continuous casting billet slabs. Among these, 780 images contain surface defects. The defects are categorized into six types: scratch (Sc), weld slag (WS), cutting opening (CO), water slag mark (WSM), slag skin (SS), and longitudinal crack (LC). 
\subsubsection{Steel Pipe} 
This dataset consists of 1,227 images of steel pipe surfaces with a fixed resolution of 728$\times$544. Among these, 554 images contain defects. The defects are divided into four types: warp, external fold, wrinkle, and scratch. 

\section{Method}
\label{sec:method}
\label{sec:formatting}

\begin{figure*}[!t]
\centering
\includegraphics[width=1.0\textwidth]{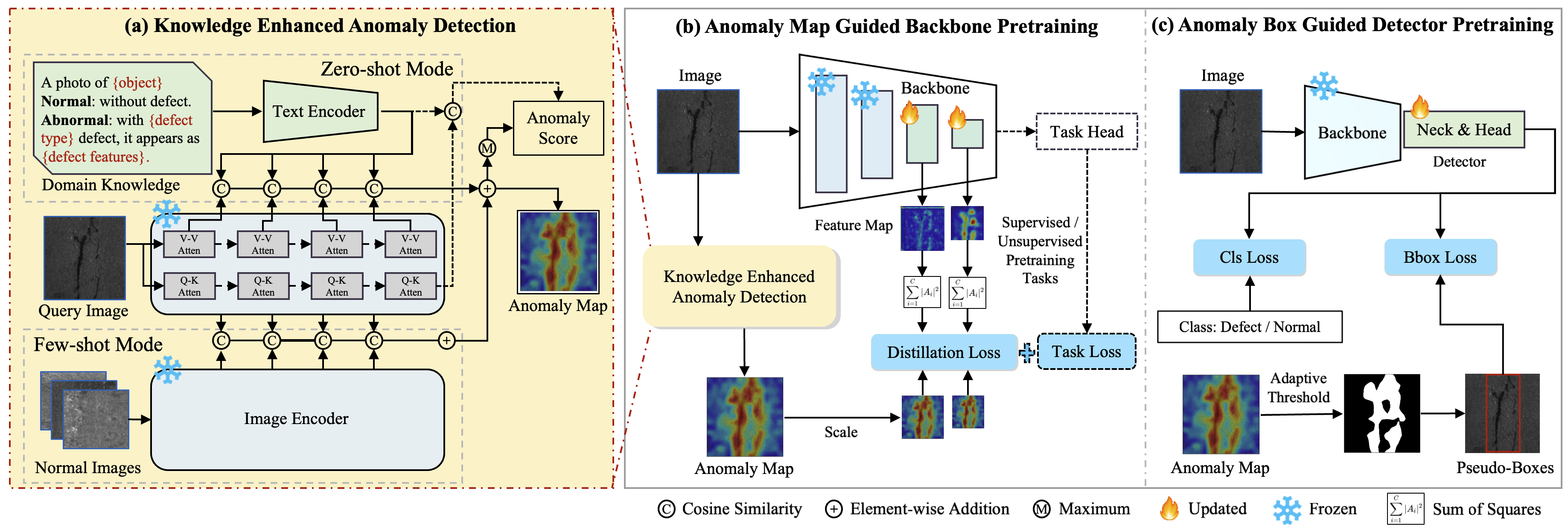}
\caption{Overview of the proposed Anomaly-Guided Self-Supervised Pretraining Framework. This framework is composed of two key phases: anomaly map guided backbone pretraining (b) and anomaly box guided detector pretraining (c). Using the Knowledge Enhanced Anomaly Detection algorithm (a), which incorporates detailed defect descriptions as prior knowledge, anomaly maps are generated. In the backbone pretraining phase, anomaly map information is transferred into the network via distillation loss. This process can be seamlessly combined with existing pretraining tasks. During the detector pretraining phase, the pretrained backbone is kept frozen, while the anomaly maps are used to generate pseudo-defect boxes. These pseudo-boxes are used for detector component pretraining within the object detection model.}
\label{fig: network}
\end{figure*}

Figure~\ref{fig: network} illustrates the overall framework of the proposed AGSSP. This framework is a two-stage pretraining process designed to be executed offline, producing a set of enhanced pretrained weights that can replace standard ImageNet weights for finetuning downstream detection models. The pretraining process involves three key components: initially, all images are input into the KeAD model to extract anomaly maps, which are stored on disk and subsequently used to guide the pretraining process. The pretraining is divided into two stages: the backbone pretraining, where knowledge distilled from the anomaly maps is incorporated into the backbone representation, and the detector pretraining, where pseudo-defect boxes derived from the anomaly maps serve as supervisory signals for detector training. Throughout the pretraining stage, the model is continuously guided to focus on abnormal characteristics in industrial defect images, including inconsistencies in color, shape, and texture. The following sections will provide a detailed explanation of the anomaly map guided backbone pretraining and anomaly box guided detector pretraining processes.

\subsection{Anomaly Map Guided Backbone Pretraining}
As illustrated in Figure~\ref{fig: network} (b), the backbone pretraining process comprises: (1) generating anomaly maps with the KeAD model to identify potential defect regions for supervision, and (2) guiding the model's high-level feature maps to align with anomaly maps during pretraining using a distillation loss, thereby improving its capacity to identify subtle anomalies. 

\subsubsection{Knowledge Enhanced Anomaly Detection}
\label{sec:kead}
Our proposed KeAD is based on the WinCLIP model~\cite{jeong2023winclip}, which extends the CLIP model~\cite{radford2021learning} for anomaly detection and segmentation. CLIP aligns images and texts in a joint embedding space, enabling zero-shot classification by measuring image–text similarity. Leveraging this property, WinCLIP performs pixel-level normal/anomaly classification via dense visual feature and normal/anomaly text embedding comparison, supporting both zero-shot and few-shot anomaly detection modes.

Specifically, in zero-shot anomaly detection, the core of this method relies on a general scoring function, $\mathcal{S}$, which quantifies the anomaly probability by comparing the similarity between a given visual feature $f_{visual} \in \mathbb{R}^d$ and corresponding normal and anomalous text representations. These representations are derived from two distinct sets of prompts: $N^+$ normal prompts, $K^+ = \{k^+_1, \dots, k^+_{N^+}\}$, and $N^-$ anomalous prompts, $K^- = \{k^-_1, \dots, k^-_{N^-}\}$. Each prompt is independently encoded into a feature vector via the CLIP text encoder, $g$. The vectors within each set are then averaged to produce the final "normal" text token $t^+\in \mathbb{R}^d$ and "anomalous" text token $t^-\in \mathbb{R}^d$. The scoring function is defined as:
\begin{equation}
\label{eq:score_func}
\mathcal{S}(f_{visual}, t^+, t^-) = \frac{\exp(\langle f_{visual}, t^- \rangle / \boldsymbol{\tau})}{\exp(\langle f_{visual}, t^+ \rangle / \boldsymbol{\tau}) + \exp(\langle f_{visual}, t^- \rangle / \boldsymbol{\tau})}
\end{equation}
where $\langle \cdot, \cdot \rangle$ denotes the cosine similarity, and $\tau > 0$ is a temperature parameter.

Notably, WinCLIP uses generic text prompts such as "Damaged" and "Defected," which lack specificity for industrial defects and lead to imprecise localization. To address this, we gathered detailed defect feature descriptions for each object category and defect type, using a structured template to generate anomaly text prompts: 
\textbf{A photo of \{object\} with \{defect type\} defect, it appears as \{defect features\}.} 
In this context, \{object\} denotes the industrial object (e.g., aluminum plate), and \{defect type\} specifies the defect type (e.g., crack). 
\textcolor{markcolor}{The \{defect features\} are constructed through a three-step process. For each defect type, we first select one representative defect image, then use GPT-4o to localize and describe the visible defect characteristics in that image, and finally refine and generalize the description using expert knowledge from industrial field practice and relevant professional references. This combination improves both descriptive precision and generalizability beyond a single sample. For each object category, the resulting prompt set covers all relevant defect types. The detailed construction process and examples are provided in the Appendix. } 

For an input image $x$, we first extract its CLS token $v \in \mathbb{R}^d$ through the ViT-based CLIP image encoder $f$. The image-level anomaly score $AS_x$ is then computed as:
\begin{equation}
\label{eq:image_score}
AS_x = \mathcal{S}(v, t^+, t^-)
\end{equation}
After the image-level score is computed, a pixel-wise anomaly map is derived by leveraging the patch tokens produced by $f$. To strengthen the semantic alignment between visual patch tokens and the text embeddings---and thus to improve spatial localization of anomalies---WinCLIP constructs enhanced dense features from the last layer of $f$ using sliding windows with varying kernel sizes. However, this strategy is memory-intensive and GPU-heavy. Our method, inspired by Clip Surgery, offers a more efficient technique. 
Clip Surgery observes that the global feature extraction process inherent in standard Q-K attention negatively impacts CLIP's localization ability. To rectify this, it proposes V-V attention, which substitutes both the Query (Q) and Key (K) components with the Value (V) component. This change keeps more local spatial information without introducing extra parameters or computations. Adopting this mechanism, we extract four sets of dense patch tokens, denoted $\mathbf{A}_{i}\in\mathbb{R}^{m\times d} (i\in\{1,2,3,4\})$, from evenly distributed intermediate layers of $f$. For every individual patch token $a_{i,j} \in A_i$ (where $j=1, \dots, m$),  a patch-level anomaly score is calculated as:
\begin{equation}
\label{eq:patch_score}
M_{i,j} = \mathcal{S}(a_{i,j}, t^+, t^-)
\end{equation}
The resulting score vector $[M_{i,1},\dots,M_{i,m}]$ is then reshaped into a 2D grid, bilinearly interpolated to the original image resolution $H\times W$, yielding the sub-anomaly map $\mathbf{M}_{i}\in\mathbb{R}^{H\times W}$. Finally, the four layer-wise maps are fused into a single zero-shot anomaly map $M_{zero}$ by averaging: 
\begin{equation}
\label{eq:final_map}
M_{zero} = \frac{1}{4} \sum_{i=1}^{4} M_i
\end{equation}

Compared to the zero-shot anomaly detection process described above, the few-shot mode introduces $k$ (typically ranging from 1 to 4) normal images as reference samples. The dense patch tokens of these reference images are extracted and stored in a memory bank $R \in \mathbb{R}^{km \times d}$. During inference, the dense patch tokens of the test image are compared with the features stored in memory to calculate similarity. Intuitively, the more dissimilar a patch is from all normal reference features, the higher its anomaly score. Accordingly, the anomaly score for each patch is defined by the maximum cosine distance between the patch and the set of reference features:

\begin{equation}
\label{formula2}
M_{few} = \mathop{min}\limits_{r \in R} \frac{1}{2} \left(1.0 - \langle A, r \rangle\right)
\end{equation}
where $M_{few}$ represents the few-shot anomaly map. The final anomaly map is obtained as $M = M_{zero} + M_{few}$. The image-level anomaly score $AS$ is then computed as the sum of the maximum value of $M$ and $AS_x$. 

\subsubsection{Backbone Pretraining}
Based on the anomaly maps obtained, we adopt an attention map-based distillation technique to transfer the anomaly cues into the model. The core idea of this distillation is to guide the backbone to focus on regions indicative of anomalies by aligning its internal feature representations with the anomaly maps. Specifically, let $A_l \in \mathbb{R}^{C_l \times H_l \times W_l}$ denote the feature map from the $l$-th layer in the backbone, we first convert it into a 2D attention map $A_l^{'} \in \mathbb{R}^{H_l \times W_l}$ by computing the sum of squares across channels:

\begin{equation}
\label{formula3}
A_l^{'} = \sum_{c=1}^{C_l} \left|A_l^c\right|^2,
\end{equation}
where $C_l, H_l, W_l$ denote the number of channels, height, and width of the feature map from the $l$-th layer, respectively.

To transfer the anomaly knowledge, an $L_2$ loss is introduced to align the multi-layer attention maps with the anomaly map. The loss function is formulated as:

\begin{equation}
\label{formula4}
L_{distill} = \sum_{l=1}^L\frac{1}{H_l \times W_l} \sum_{i=1}^{H_l} \sum_{j=1}^{W_l} \left( A_l^{'}(i, j) - M_l(i, j) \right)^2
\end{equation}
where $L$ is the total number of layers considered, $M_l \in \mathbb{R}^{H_l \times W_l}$ is the corresponding anomaly map, resized to match the spatial dimensions of $A_l^{'}$. $(i, j)$ indexes the spatial location in the attention or anomaly map. 
Crucially, the target anomaly maps $M_l$ are soft confidence signals, with pixel values indicating the likelihood of anomaly presence. By regressing these maps, the network learns to focus on defect-like patterns while suppressing background in low-confidence regions. This soft-supervision mechanism enhances robustness to noise in the large-scale unlabeled dataset. 

As shown in Figure~\ref{fig: network} (b), the distillation loss can be combined with any other pretraining task losses $L_{task}$, providing complementary advantages. We integrate it into three pretraining task categories: supervised pretraining for image classification, self-supervised pretraining via contrastive learning (e.g., MoCov3~\cite{chen2021empirical}), and image reconstruction tasks like SimMIM~\cite{xie2022simmim}. This integration strengthens the model's capacity to learn robust and transferable representations. The formulation incorporating distillation loss with existing task losses is given by:

\begin{equation}
\label{formula5}
L_{total} = L_{distill} + \lambda L_{task}
\end{equation}
where $\lambda$ is a hyperparameter that adjusts the weights of the two loss functions, balancing their relative impact. To simplify the training process, $\lambda$ is determined based on the ratio of these two loss functions values during the first iteration: $\lambda = L_{distill}^{0} / L_{task}^{0}$. This method balances the initial magnitudes of these two loss functions for stable training.

It should be noted that only high-level features are used for the distillation loss, while lower-level features are sourced from publicly available pretrained models trained on ImageNet. This design stems from the insight that lower-level feature maps generally capture universal, data-agnostic patterns transferrable across various domains. Conversely, high-level feature maps emphasize task-specific semantic representations that closely match the pretraining goals. This method allows the model to efficiently enhance semantic features unique to anomalies, while simultaneously utilizing the adaptable, generic lower-level features of ImageNet pretrained models.

Additionally, while anomaly maps originate from ViT models, the attention map distillation technique enables their transfer to architectures like ResNet~\cite{he2016deep} and CSPDarknet~\cite{wang2021scaled}. Our empirical findings suggest that using cosine distance to align anomaly and attention maps enhances transferability to non-ViT backbones, as it emphasizes relative feature space relationships over absolute values. 
Therefore, we employ $L_2$ loss for transfers within ViT architectures and use cosine distance-based loss for transfers between different architectures.

\subsection{Anomaly Box Guided Detector Pretraining}
\label{sec:agdp}
To mitigate the performance degradation caused by the random initialization of the detector during finetuning, we design a detector pretraining method based on the pretrained backbone, aligned with the downstream defect detection task. In this pipeline, pseudo-defect boxes are generated from anomaly maps and subsequently used to supervise the detector training. 

The pipeline for generating pseudo-defect boxes and pretraining the detector is shown in Figure~\ref{fig: network} (c). First, an image $x$ is classified according to its anomaly score $AS$: if $AS < 0.5$, it is labeled as normal; otherwise, it is labeled as defect. For defective images, pseudo-defect boxes are then generated. To this end, we design a dynamic thresholding strategy that adapts to each object category, ensuring robust binarization across diverse data distributions. Specifically, given an anomaly map $M(i, j)$, a category-specific threshold $T_k$ is computed to produce a binary mask $BM(i, j)$. For the large-scale industrial dataset with $n$ object categories $D = \{ D_1, D_2, \dots, D_n\}$, with $D_k$ representing anomaly maps for category $k$, $T_k$ is computed as: 

\begin{equation}
\label{formula6}
T_k = \mu_{max}(D_k) - \delta
\end{equation}
Here, $\mu_{max}(D_k)$ is the mean of the maximum anomaly values across all anomaly maps in category $k$:

\begin{equation}
\label{formula7}
\mu_{max}(D_k) = \frac{1}{\left|D_k\right|}\sum_{M \in D_k}max(M(i, j))
\end{equation} 
The constant $\delta$ is set to 0.1 based on our ablation study to adjust the threshold slightly below $\mu_{max}(D_k)$, thereby enhancing the separation between normal and defective regions. This dynamic design is motivated by the observation that different datasets exhibit distinct anomaly-value distributions, making a fixed threshold suboptimal; moreover, tuning an optimal value for each dataset is labor-intensive, whereas our approach adapts thresholds automatically in a data-driven manner.
The binary mask $BM(i, j)$ for each defective image is then defined as: 
\begin{equation}
\begin{aligned}
\label{formula8}
BM(i, j) = 
& \begin{cases} 1 & \mbox{if }M(i, j) \geq T_k \\
0 & \mbox{otherwise}
\end{cases}
\end{aligned}
\end{equation}
After obtaining the binary mask, the next step is to convert it into defect-level bounding boxes suitable for detector pretraining. Connected regions of anomalies are first extracted from $BM(i, j)$ using the connectedComponentsWithStats function in OpenCV. For each connected component $C_i$, its bounding box $B_i$ is computed as the smallest bounding rectangle that encloses the component. 
Among all bounding boxes, the top 10 bounding boxes with the highest anomaly scores are retained as labels, which align with the characteristics of the metallic surface defect detection scenario. To further reduce redundancy, Non-Maximum Suppression (NMS) is applied to merge overlapping or closely located boxes. This process effectively converts anomaly maps into defect boxes, which serve as pseudo-labels for pretraining object detection models. 

\section{Experiments}
\label{sec:experiments}

\subsection{Experimental Setup}
All pretrained models were trained on two NVIDIA A100 GPUs (40GB), while finetuned models were trained on an NVIDIA GeForce RTX 3090 GPU (24GB).

\subsubsection{Datasets}
In addition to using open-source pretrained models from ImageNet and COCO, we conducted additional pretraining on our large-scale industrial dataset, followed by finetuning and validation on two labeled metallic surface defect datasets: Casting Billet and Steel Pipe. Details of these datasets are provided in Section~\ref{sec:datasets}. The training and validation sets for Casting Billet and Steel Pipe were split in a 1:1 ratio. We also generated Casting Billet-Mini50, creating an extreme few-shot setting by randomly selecting 50 samples for training. 
Model generalizability was further assessed using the open-source GC10-DET dataset~\cite{lv2020deep}, which contains 2,300 images spanning 10 types of steel surface defects. We randomly selected 500 samples from the training set to create GC10-Mini, with a training set size that aligns with the other two datasets, representing data-limited scenarios. Additionally, we validated the model on a non-metallic surface dataset, the Fabric Defects Detection Dataset (FD-Dataset)~\cite{neamah2024fabric}, which contains 540 training images and 107 validation images, to examine cross-domain performance. 

To comprehensively evaluate the effectiveness of our proposed KeAD module, we performed comparative and ablation studies on our collected metallic surface datasets (Casting Billet and Steel Pipe), and further validated its performance and generalizability on two widely recognized industrial anomaly detection benchmarks (MVTec AD~\cite{bergmann2019mvtec} and VisA~\cite{zou2022spot}). 

\subsubsection{Implementation Details}

\textcolor{markcolor}{For anomaly-map generation, OpenCLIP ViT-H-14-378-quickgelu was used as the encoder, together with its corresponding text tokenizer, in a few-shot setting with four normal reference samples randomly selected and then fixed for each category. Specifically, KeAD computed anomaly maps based on the detailed text prompt templates described in Section~\ref{sec:kead}. After standard min-max normalization, these maps served directly as soft anomaly-confidence maps for backbone pretraining. For detector pretraining, they were further processed into pseudo boxes via category-wise adaptive thresholding, connected-component analysis, and NMS, as described in Section~\ref{sec:agdp}.} 

For both pretraining and finetuning, our implementation was built on MMDetection with YOLOv8, using four backbones: CSPDarknet, ResNet50, WideResNet50~\cite{zagoruyko2016wide}, and Swin-Base. Beyond conventional supervised pretraining, we also evaluated representative self-supervised methods tailored to different architectures, namely MoCoV3~\cite{chen2021empirical} for the CNN-based ResNet and SimMIM~\cite{xie2022simmim} for the Transformer-based Swin-Base. 
Training configurations are summarized in Table~\ref{tab:training_configs}. Optimizer, scheduler, and data augmentation followed the settings of the corresponding model literature. The batch size was determined by the GPU capacity and the learning rates were adjusted proportionally. Backbone pretraining epochs were selected through ablation studies, while a small number of epochs were used for detector pretraining to avoid overfitting. The finetuning process was limited to 500 epochs, with best results determined based on evaluation metrics. All images were processed uniformly. For pretraining, images were uniformly resized to 224$\times$224, matching ImageNet-pretrained models. For finetuning, they were resized to 640$\times$640, aligned with YOLOv8 model. \textcolor{markcolor}{For practical reference, the approximate GPU hours (GPUh) required for each pretraining stage are detailed alongside the training configurations in Table~\ref{tab:training_configs}.}

\begin{table}[!t]
\centering
\fontsize{8}{10}\selectfont
\setlength{\tabcolsep}{0.8mm} 

\begin{tabular}{llcccccc} 
\toprule
\textbf{Backbone} & \textbf{Method} & \textbf{Stage} & \textbf{BS} & \textbf{Optimizer} & \textbf{LR} & \textbf{Epochs} & \textbf{\textcolor{markcolor}{GPUh}} \\ 
\midrule

\multirow{3}{*}{\textbf{CSPDarknet}} & \multirow{3}{*}{Cls} 
 & BP & 1024 & SGD & 0.01 & 200 & 22.1 \\
 & & DP & 256 & SGD & 0.01 & 10 & 2.70 \\
 & & FT & 32 & SGD & 0.0025 & 500 & -- \\ 
\midrule

\multirow{3}{*}{\makecell[l]{\textbf{ResNet50}/\\ \textbf{WideResNet50}}} & \multirow{3}{*}{Cls/MoCov3} 
 & BP & 512 & SGD & 0.01 & 200 & 40.7 \\
 & & DP & 128 & SGD & 0.0025 & 10 & 5.25 \\
 & & FT & 32 & SGD & 0.0025 & 500 & -- \\ 
\midrule

\multirow{3}{*}{\textbf{Swin-Base}} & \multirow{3}{*}{Cls/SimMIM} 
 & BP & 512 & AdamW & 0.0001 & 200 & 44.0 \\
 & & DP & 128 & SGD & 0.0025 & 10 & 6.22 \\
 & & FT & 14 & SGD & 0.0025 & 500 & -- \\ 
\bottomrule
\end{tabular}

\caption{Training configurations. BS: Batch Size; LR: Learning Rate; BP: Backbone Pretraining; DP: Detector Pretraining; FT: Finetuning; Cls: Classification; \textcolor{markcolor}{GPUh: Approximate GPU hours on 1$\times$A100.}}
\label{tab:training_configs}
\end{table}

\subsubsection{Evaluation Metrics}
Common object detection metrics, including mAP@0.5 (denoted as mAP@.5 in tables) and mAP@0.5:0.95 (denoted as mAP@.5:95), were utilized for defect detection evaluation. For anomaly detection, the Area Under the Receiver Operating Characteristic (AUROC) metric of image and pixel levels were employed as standard criteria. 

\begin{table*}[!t]
\centering
\fontsize{6}{9}\selectfont
\setlength{\tabcolsep}{0.4mm}{
\begin{tabular}{lllllllllll}
\toprule
\multicolumn{1}{c}{} & \multicolumn{1}{c}{} & \multicolumn{1}{c}{Pretraining} & \multicolumn{2}{c}{Casting Billet} & \multicolumn{2}{c}{Steel Pipe}  & \multicolumn{2}{c}{GC10-Mini} & \multicolumn{2}{c}{GC10-Det}                                    \\ \cline{4-11} 
\multicolumn{1}{c}{\multirow{-2}{*}{Backbone}}   & \multicolumn{1}{c}{\multirow{-2}{*}{Method}} & \multicolumn{1}{c}{Dataset} & mAP@.5$\uparrow$   & \multicolumn{1}{c}{mAP@.5:95$\uparrow$} & mAP@.5$\uparrow$ & \multicolumn{1}{c}{mAP@.5:95$\uparrow$} & mAP@.5$\uparrow$ & \multicolumn{1}{c}{mAP@.5:95$\uparrow$} & mAP@.5$\uparrow$ & \multicolumn{1}{c}{mAP@.5:95$\uparrow$} \\ \midrule

& - & - & 74.2 & 37.1 & 68.8 & 35.5 & 48.1 & 22.7 & 61.6 & 30.0 \\
& Detection & COCO & 75.1 & 43.9 & 72.7 & 41.6 & \textbf{59.3} & 28.1 & 63.8 & 31.5 \\
& Cls & ImageNet & 71.6 & 33.5 & 74.3 & 38.7 & 49.1 & 22.9 & 62.6 & 29.4 \\
& Cls & Industrial & 72.8\textcolor{agreen}{\textcolor{agreen}{(+1.2)}} & 34.7\textcolor{agreen}{\textcolor{agreen}{(+1.2)}} & 73.3\textcolor{ashgrey}{\textcolor{ashgrey}{(-1.0)}} & 38.2\textcolor{ashgrey}{(-0.5)} & 49.6\textcolor{agreen}{(+0.5)} & 23.4\textcolor{agreen}{(+0.5)} & 62.3\textcolor{ashgrey}{(-0.3)} & 29.1\textcolor{ashgrey}{(-0.3)} \\
& +AGBP & Industrial & 74.9\textcolor{agreen}{(+3.3)} & 37.5\textcolor{agreen}{(+4.0)} & 74.2\textcolor{ashgrey}{(-0.1)} & 39.7\textcolor{agreen}{(+1.0)} & 49.7\textcolor{agreen}{(+0.6)} & 24.0\textcolor{agreen}{(+1.1)} & 62.1\textcolor{ashgrey}{(-0.5)} & 29.5\textcolor{agreen}{(+0.1)} \\
\multirow{-6}{*}{CSPDarknet} & \textbf{+AGSSP} & \textbf{Industrial} & \textbf{77.2\textcolor{agreen}{(+5.6)}} & \textbf{44.9\textcolor{agreen}{(+11.4)}} & \textbf{76.4\textcolor{agreen}{(+2.1)}} & \textbf{42.7\textcolor{agreen}{(+4.0)}} & 58.9\textcolor{agreen}{(+9.8)} & \textbf{28.3\textcolor{agreen}{(+5.4)}} & \textbf{66.6\textcolor{agreen}{(+4.0)}} & \textbf{32.3\textcolor{agreen}{(+2.9)}} \\ \midrule
& - & - & 56.7 & 27.0 & 37.2 & 16.3 & 43.8 & 20.6 & 54.1 & 25.8 \\
& Cls & ImageNet & 69.1 & 32.3 & 73.8 & 38.9 & 50.4 & 23.3 & 61.4 & 29.5 \\
& Cls & Industrial & 69.1\textcolor{agreen}{(+0.0)} & 33.6\textcolor{agreen}{(+1.3)} & 73.2\textcolor{ashgrey}{(-0.6)} & 39.8\textcolor{agreen}{(+0.9)} & 49.3\textcolor{ashgrey}{(-1.1)} & 23.7\textcolor{agreen}{(+0.4)} & 60.1\textcolor{ashgrey}{(-1.3)} & 29.1\textcolor{ashgrey}{(-0.4)} \\
& +AGBP & Industrial & 72.0\textcolor{agreen}{(+2.9)} & 36.1\textcolor{agreen}{(+3.8)} & 74.9\textcolor{agreen}{(+1.1)} & 40.6\textcolor{agreen}{(+1.7)} & 49.9\textcolor{ashgrey}{(-0.5)} & 23.2\textcolor{ashgrey}{(-0.1)} & 62.8\textcolor{agreen}{(+1.4)} & 29.0\textcolor{ashgrey}{(-0.5)} \\
& \textbf{+AGSSP} & \textbf{Industrial} & \textbf{74.5\textcolor{agreen}{(+5.4)}} & \textbf{42.5\textcolor{agreen}{(+10.2)}} & \textbf{76.7\textcolor{agreen}{(+2.9)}} & \textbf{42.3\textcolor{agreen}{(+3.4)}} & \textbf{59.1\textcolor{agreen}{(+8.7)}} & \textbf{28.3\textcolor{agreen}{(+5.0)}} & \textbf{65.3\textcolor{agreen}{(+3.9)}} & \textbf{31.5\textcolor{agreen}{(+2.0)}} \\ \cline{2-11}

& DINOv1 & ImageNet & 67.8 & 34.5 & 69.1 & 34.8 & 50.6 & 23.5 & 60.2 & 27.7 \\ \cline{2-3}

& MoCov3 & ImageNet & 70.2 & 33.9 & 70.7 & 37.1 & 48.3 & 22.7 & 59.9 & 28.4 \\
& MoCov3 & Industrial & 71.3\textcolor{agreen}{(+1.1)} & 36.8\textcolor{agreen}{(+2.9)} & 70.6\textcolor{ashgrey}{(-0.1)} & 37.0\textcolor{ashgrey}{(-0.1)} & 48.4\textcolor{agreen}{(+0.1)} & 22.5\textcolor{ashgrey}{(-0.2)} & 61.2\textcolor{agreen}{(+1.3)} & 27.2\textcolor{ashgrey}{(-1.2)} \\
\multirow{-9}{*}{ResNet50} & +AGBP & Industrial & 71.4\textcolor{agreen}{(+1.2)} & 33.1\textcolor{ashgrey}{(-0.8)} & 71.0\textcolor{agreen}{(+0.3)} & 36.6\textcolor{ashgrey}{(-0.5)} & 49.2\textcolor{agreen}{(+0.9)} & 23.9\textcolor{agreen}{(+1.2)} & 58.6\textcolor{ashgrey}{(-1.3)} & 27.7\textcolor{ashgrey}{(-0.7)} \\ \midrule

& Cls & ImageNet & 71.5 & 34.6 & 69.9 & 35.8 & 50.5 & 23.5 & 61.8 & 29.9 \\
& +AGBP & Industrial & 72.5\textcolor{agreen}{(+1.0)} & 37.8\textcolor{agreen}{(+3.2)} & 71.7\textcolor{agreen}{(+1.8)} & 36.8\textcolor{agreen}{(+1.0)} & 52.1\textcolor{agreen}{(+1.6)} & 24.3\textcolor{agreen}{(+0.8)} & 62.0\textcolor{agreen}{(+0.2)} & 29.8\textcolor{ashgrey}{(-0.1)} \\
\multirow{-3}{*}{WideResNet50} & \textbf{+AGSSP} & \textbf{Industrial} & \textbf{73.5\textcolor{agreen}{(+2.0)}} & \textbf{42.5\textcolor{agreen}{(+7.9)}} & \textbf{74.2\textcolor{agreen}{(+4.3)}} & \textbf{40.4\textcolor{agreen}{(+4.6)}} & 60.3\textcolor{agreen}{(+9.8)} & \textbf{28.9\textcolor{agreen}{(+5.4)}} & \textbf{63.7\textcolor{agreen}{(+1.9)}} & \textbf{32.5\textcolor{agreen}{(+2.6)}} \\ \midrule

& - & - & 61.1 & 30.1 & 55.7 & 27.7 & 44.7 & 22.3 & 54.8 & 26.5 \\
& Cls & ImageNet & 75.7 & 37.7 & 78.0 & 39.0 & 49.7 & 23.8 & 63.1 & 31.8 \\
& Cls & Industrial & 73.3\textcolor{ashgrey}{(-2.4)} & 36.0\textcolor{ashgrey}{(-1.7)} & 74.9\textcolor{ashgrey}{(-3.1)} & 37.1\textcolor{ashgrey}{(-1.9)} & 51.6\textcolor{agreen}{(+1.9)} & 24.2\textcolor{agreen}{(+0.4)} & 60.0\textcolor{ashgrey}{(-3.1)} & 28.2\textcolor{ashgrey}{(-3.6)} \\
& +AGBP & Industrial & 76.8\textcolor{agreen}{(+1.1)} & 39.0\textcolor{agreen}{(+1.3)} & 78.2\textcolor{agreen}{(+0.2)} & 39.9\textcolor{agreen}{(+0.9)} & 49.8\textcolor{agreen}{(+0.1)} & 23.7\textcolor{ashgrey}{(-0.1)} & 61.2\textcolor{ashgrey}{(-1.9)} & 30.3\textcolor{ashgrey}{(-1.5)} \\
& \textbf{+AGSSP} & \textbf{Industrial} & 77.7\textcolor{agreen}{(+2.0)} & 45.0\textcolor{agreen}{(+7.3)} & 78.7\textcolor{agreen}{(+0.7)} & \underline{\textbf{44.0\textcolor{agreen}{(+5.0)}}} & \underline{\textbf{59.7\textcolor{agreen}{(+10.0)}}} & 28.1\textcolor{agreen}{(+4.3)} & 63.5\textcolor{agreen}{(+0.4)} & 32.0\textcolor{agreen}{(+0.2)} \\ \cline{2-11}
& SimMIM & ImageNet & 78.3 & 44.3 & 77.5 & 40.7 & 51.0 & 23.8 & 63.4 & 31.0 \\
& SimMIM & Industrial & 76.5\textcolor{ashgrey}{(-1.8)} & 41.5\textcolor{ashgrey}{(-2.8)} & 77.5\textcolor{agreen}{(+0.0)} & 40.1\textcolor{ashgrey}{(-0.6)} & 50.0\textcolor{ashgrey}{(-1.0)} & 23.7\textcolor{ashgrey}{(-0.1)} & 60.9\textcolor{ashgrey}{(-2.5)} & 29.7\textcolor{ashgrey}{(-1.3)} \\
& +AGBP & Industrial & \underline{\textbf{79.8\textcolor{agreen}{(+1.5)}}} & 44.3\textcolor{agreen}{(+0.0)} & \underline{\textbf{81.4\textcolor{agreen}{(+3.9)}}} & 43.7\textcolor{agreen}{(+3.0)} & 52.1\textcolor{agreen}{(+1.1)} & 24.7\textcolor{agreen}{(+0.9)} & 62.1\textcolor{ashgrey}{(-1.3)} & 31.4\textcolor{agreen}{(+0.4)} \\
\multirow{-9}{*}{Swin-Base} & \textbf{+AGSSP} & \textbf{Industrial} & 78.8\textcolor{agreen}{(+0.5)} & \underline{\textbf{48.4\textcolor{agreen}{(+4.1)}}} & 79.0\textcolor{agreen}{(+1.5)} & 43.7\textcolor{agreen}{(+3.0)} & 58.8\textcolor{agreen}{(+7.8)} & \underline{\textbf{29.2\textcolor{agreen}{(+5.4)}}} & \underline{\textbf{67.7\textcolor{agreen}{(+4.3)}}} & \underline{\textbf{34.0\textcolor{agreen}{(+3.0)}}} \\ 
\bottomrule
\end{tabular}
}
\caption{Comparison of finetuning results for various pretrained model configurations on four metallic defect detection datasets. The configurations differ in terms of backbone architectures, pretraining methods, and pretraining datasets. The \textbf{bold} numbers indicate the best performance within the same backbone. The \underline{underlined} numbers represent the overall highest performance across all configurations. Here, AGSSP means AGBP plus AGDP. - means training from scratch without any pretrained models.}
\label{tab:main_results}
\end{table*}

\subsection{Results Comparison and Analysis}
\subsubsection{Main Results} 
Table~\ref{tab:main_results} summarizes the evaluation results of pretraining techniques on four datasets: Casting Billet, Steel Pipe, GC10-Mini, and GC10-DET. In general, four distinct pretraining modes were analyzed in four backbone architectures. Specifically, Mode 1 employs ImageNet-pretrained models. Mode 2 continues training these models with our large-scale industrial dataset. Mode 3 (+AGBP) incorporates an anomaly map distillation loss during the backbone training of Mode 1's models. Mode 4 (+AGSSP) builds upon Mode 3 by additionally pretraining the detector. The effectiveness of supervised pretraining method for image classification was evaluated across all backbones, where the 61 object categories of the industrial dataset were taken as its classification labels. We further explored other pretraining methods: MoCov3 pretrained ResNet50, SimMIM pretrained Swin-Base and COCO pretrained CSPDarknet. Table~\ref{tab:main_results} also includes the performance of training from scratch, denoted by -.

An in-depth analysis of the results presented in Table~\ref{tab:main_results} uncovers several important findings. 
First, Mode 2 underperforms compared to Mode 1 in most cases, indicating that leveraging current pretraining techniques on large industrial datasets leads to suboptimal performance. This could stem from the restricted size and variation of industrial datasets or the mismatch between existing pretraining pretext tasks and industrial contexts. 
Second, the performance of Mode 3 generally exceeds that of Modes 2 and 1, except in the GC10-Det context where Mode 3 performs comparably to Mode 2. This demonstrates the efficacy of the anomaly map-guided pretraining method for industrial applications, especially with limited data availability.
Third, Mode 4 consistently achieves the best or near-best performance across various datasets and metrics, notably surpassing both ImageNet and COCO-based pretrained models. This highlights the effectiveness of leveraging anomaly maps to produce pseudo-defect labels for detector pretraining on industrial datasets. Notably, Mode 4 yields up to a 4.3\% enhancement in mAP on the GC10-Det, which is relatively data-abundant, and over a 10\% increase on data-scarce datasets like GC10-Mini and Casting Billet. 
We further evaluated our AGSSP framework on WideResNet50, an empirically effective variant of ResNet50, and observed consistent improvements, with gains of +7.9\% and +4.6\% in mAP@0.5:0.95 on Casting Billet and Steel Pipe, respectively, demonstrating its robustness on high-capacity backbones. Meanwhile, MoCov3 and DINOv1~\cite{caron2021emerging} pretrained models fall short of their supervised counterparts across all datasets, and gain little from AGBP, highlighting their limited suitability for metallic defect detection. 

\begin{table}[!t]
\centering
\fontsize{8}{10}\selectfont
\setlength{\tabcolsep}{0.8mm}{
\begin{tabular}{llllll}
\toprule
\multirow{2}{*}{Backbone} & \multirow{2}{*}{Pretraining} & \multicolumn{2}{c}{Casting Billet-Mini50} & \multicolumn{2}{c}{FD-Dataset} \\ \cline{3-6}
 & & \multicolumn{1}{c}{mAP@.5} & mAP@.5:95 & \multicolumn{1}{c}{mAP@.5} & \multicolumn{1}{c}{mAP@.5:95}  \\
\midrule
\multirow{3}{*}{CSPDarknet} 
 & COCO & 52.7 & 29.3 & 86.2 & 49.4 \\
 & ImageNet & 23.6 & 7.90 & 86.1 & 47.2 \\
 & \textbf{AGSSP} & \textbf{55.8\textcolor{agreen}{(+32.2)}} & \textbf{29.5\textcolor{agreen}{(+21.6)}} & \textbf{89.3\textcolor{agreen}{(+3.2)}} & \textbf{50.7\textcolor{agreen}{(+3.5)}} \\
\midrule
\multirow{2}{*}{ResNet50} 
 & ImageNet & 30.8 & 11.2 & 87.5 & 50.5 \\
 & \textbf{AGSSP} & \textbf{50.9\textcolor{agreen}{(+20.1)}} & \textbf{25.4\textcolor{agreen}{(+14.2)}} & \textbf{89.0\textcolor{agreen}{(+1.5)}} & \textbf{51.7\textcolor{agreen}{(+1.2)}} \\
\midrule
\multirow{2}{*}{Swin-Base} 
 & SimMIM & 15.3 & 5.30 & 91.1 & 52.2 \\
 & \textbf{AGSSP} & \underline{\textbf{56.3\textcolor{agreen}{(+41.0)}}} & \underline{\textbf{29.8\textcolor{agreen}{(+24.5)}}} & \underline{\textbf{93.8\textcolor{agreen}{(+2.7)}}} & \underline{\textbf{57.4\textcolor{agreen}{(+5.2)}}} \\
\bottomrule
\end{tabular}}
\caption{Comparison of finetuning results on Casting Billet-Mini50 and FD-Dataset.}
\label{tab:mini}
\end{table}

Table~\ref{tab:mini} further demonstrates AGSSP's superior performance under two challenging conditions. In an extreme few-shot learning scenario (Casting Billet-Mini50), the framework achieves a remarkable gain of up to +41\% in mAP@0.5 and +24.5\% in mAP@0.5 :0.95. Furthermore, when tested for cross-domain generalization on a non-metallic dataset (FD-Dataset), AGSSP robustly improves performance, showing gains of up to +3.2\% in mAP@0.5 and +5.2\% in mAP@0.5:0.95. These results highlight AGSSP's ability to learn generalizable representations.

\begin{table}[!t]
\centering
\fontsize{8}{10}\selectfont
\setlength{\tabcolsep}{0.9mm}{

\begin{tabular}{c|c|c|c|c|cc|cc}
\toprule
\multirow{2}{*}{Model} & \multirow{2}{*}{Encoder} & \multirow{2}{*}{Mode} & \multirow{2}{*}{DF} & \multirow{2}{*}{V-V} & \multicolumn{2}{c|}{Casting Billet} & \multicolumn{2}{c}{Steel Pipe} \\ \cline{6-9}
& &  &  & & \multicolumn{1}{c}{Image$\uparrow$} & \multicolumn{1}{c|}{Pixel$\uparrow$} & \multicolumn{1}{c}{Image$\uparrow$} & \multicolumn{1}{c}{Pixel$\uparrow$} \\ \midrule
\multirow{2}{*}{WinCLIP} & \multirow{2}{*}{Base} & Zero & \XSolidBrush & \XSolidBrush & 65.8 & 63.3 & 52.6 & 67.8 \\ \cline{3-9}
& & Few & \XSolidBrush & \XSolidBrush & 72.8 & 73.0 & 58.3 & \textbf{77.3} \\ \hline

\multirow{2}{*}{April-GAN} & \multirow{2}{*}{Large} & Zero & - & - & 84.5 & 77.2 & 59.0 & 77.3 \\ \cline{3-9}
& & Few & - & - & 79.8 & 77.1 & 60.4 & \underline{\textbf{82.5}} \\ \hline

AnomalyCLIP & Large & Zero & - & - & 79.9 & 81.9 & 65.7 & 78.1 \\ \hline
\textcolor{markcolor}{AdaCLIP} & Large & Zero & - & - & 59.8 & 71.4 & 54.8 & 76.8 \\ \hline
AA-CLIP & Large & Zero & - & - & 81.3 & \underline{\textbf{83.9}} & 55.8 & 78.4 \\ \hline

\multirow{7}{*}{KeAD} & \multirow{3}{*}{Base} & Few & \Checkmark & \XSolidBrush & 77.4 & 73.8 & \textbf{61.1} & 75.9 \\ \cline{3-9}
& & Few & \XSolidBrush & \Checkmark & 79.4 & 76.4 & 54.4 & 70.0 \\ \cline{3-9}
& & Few & \Checkmark & \Checkmark & \textbf{82.7} & \textbf{80.7} & 58.9 & 75.5 \\ \cline{2-9}

 & \multirow{2}{*}{Large} & Few & \XSolidBrush & \Checkmark & 88.9 & 76.7 & 62.4 & 75.9 \\ \cline{3-9}
 & & Few & \Checkmark & \Checkmark & \textbf{90.2} & 77.9 & \textbf{65.1} & 77.7 \\ \cline{2-9}
 
 & \multirow{2}{*}{Huge} & Few & \XSolidBrush & \Checkmark & 92.3 & 76.5 & 65.1 & 76.7 \\ \cline{3-9}
 & & Few & \Checkmark & \Checkmark & \underline{\textbf{93.5}} & \textbf{77.7} & \underline{\textbf{69.0}} & \textbf{81.0} \\
\bottomrule
\end{tabular}
}
\caption{Quantitative results and ablation study of the knowledge enhanced anomaly detection model. The DF means Defect Feature. The V-V means V-V Attention.}
\label{tab:anomaly_detection}
\end{table}

\begin{table}[t]
    \centering
    \fontsize{8}{10}\selectfont
    \setlength{\tabcolsep}{0.9mm}{
    
    \begin{tabular}{c|c|c| cc|cc}
        \toprule
        \multirow{2}{*}{Model} & \multirow{2}{*}{Encoder} & \multirow{2}{*}{Mode} & \multicolumn{2}{c|}{MVTec AD} & \multicolumn{2}{c}{VisA} \\
        \cline{4-7}
         & & & Image$\uparrow$ & Pixel$\uparrow$ & Image$\uparrow$ & Pixel$\uparrow$ \\
        \midrule
        \multirow{2}{*}{WinCLIP} & \multirow{2}{*}{Base} & Zero & 91.8 & 85.1 & 78.1 & 79.6 \\ \cline{3-7}
         & & Few & 95.2 & \textbf{96.2} & 87.3 & 97.2 \\
        \hline
        \multirow{2}{*}{April-GAN} & \multirow{2}{*}{Large} & Zero & 86.1 & 87.6 & 78.0 & 94.2 \\ \cline{3-7}
         & & Few & 92.8 & 95.9 & \underline{\textbf{92.6}} & 96.2 \\
        \hline
        AnomalyCLIP & Large & Zero & 91.5 & 91.1 & 82.0 & \textbf{95.5} \\
        \hline
        \textcolor{markcolor}{AdaCLIP} & Large & Zero & 89.2 & 88.7 & \textbf{85.8} & \textbf{95.5} \\
        \hline
        AA-CLIP & Large & Zero & 90.5 & \textbf{91.9} & 84.6 & \textbf{95.5} \\
        \hline
        \multirow{2}{*}{KeAD} & \multirow{2}{*}{Huge} & Zero & \textbf{92.5} & 90.6 & 84.5 & 94.2 \\ \cline{3-7}
         & & Few & \underline{\textbf{96.6}} & \underline{\textbf{96.2}} & 91.5 & \underline{\textbf{97.9}} \\
        \bottomrule
    \end{tabular}
    }
    \caption{Quantitative comparison of zero-shot and few-shot anomaly detection performance on widely used public benchmarks (MVTec AD and VisA). \textbf{Bold} indicates the best performance within each mode.}
    \label{tab:mvtec_visa_comparison}
\end{table}

\subsubsection{Anomaly Detection Results}
Table~\ref{tab:anomaly_detection} presents the performance of the proposed KeAD method on the Casting Billet and Steel Pipe datasets. In the Encoder column, "Base", "Large" and "Huge" refer to OpenCLIP’s ViT-B-16-plus-240-laion400m\_e32, ViT-L-14-336-openai and ViT-H-14-378-quickgelu models, respectively. \textcolor{markcolor}{The Defect Feature (DF) column indicates whether detailed defect-description prompts are used; w/o DF means using only generic anomaly prompts such as ``damaged'' and ``defected''.} 
Compared with the WinCLIP model, the results indicate that our proposed KeAD method significantly enhances performance by integrating two key optimization strategies. Specifically, incorporating detailed defect description information reliably boosts performance in both datasets and remains effective across various encoder architectures. This validates that introducing domain-specific textual priors is a robust way to enhance anomaly localization. In terms of visual feature extraction, the introduction of V-V attention improves performance on the Casting Billet while showing reduced efficacy on the Steel Pipe. This suggests that the V-V attention mechanism efficiently captures anomalies in the Casting Billet, whereas WinCLIP's sliding-window approach is more adept at anomaly detection in the Steel Pipe, emphasizing the distinct characteristics of each dataset. However, it is worth noting that KeAD achieves substantial computational efficiency gains by adopting V-V attention. It processes the Steel Pipe in just 5 minutes compared to WinCLIP's 52 minutes on an NVIDIA 3090 GPU, while avoiding WinCLIP's memory constraints with the Huge model.

To further demonstrate the versatility of KeAD, we extended our evaluation to the standard MVTec AD and VisA benchmarks. As detailed in Tables~\ref{tab:mvtec_visa_comparison} and \ref{tab:anomaly_detection}, we compared our method against WinCLIP and recent state-of-the-art (SOTA) approaches, including April-GAN, AnomalyCLIP, \textcolor{markcolor}{AdaCLIP}, and AA-CLIP. Comparison of results across the two tables reveals a notable performance divergence. While methods utilizing auxiliary training datasets excel on public benchmarks, they suffer significant drops on our Casting Billet and Steel Pipe datasets. This is exemplified by \textcolor{markcolor}{AdaCLIP}, which achieves competitive results on VisA but lags behind KeAD by a wide margin on Steel Pipe (54.8\% vs. 69.0\% in image-level metrics). The large performance gap indicates that priors learned from general auxiliary datasets struggle to generalize to the complex textures of metallic surfaces. In contrast, KeAD delivers high performance across both domain-specific and general industrial datasets.

\begin{figure}[!t]
\centering
\includegraphics[width=0.9\columnwidth]{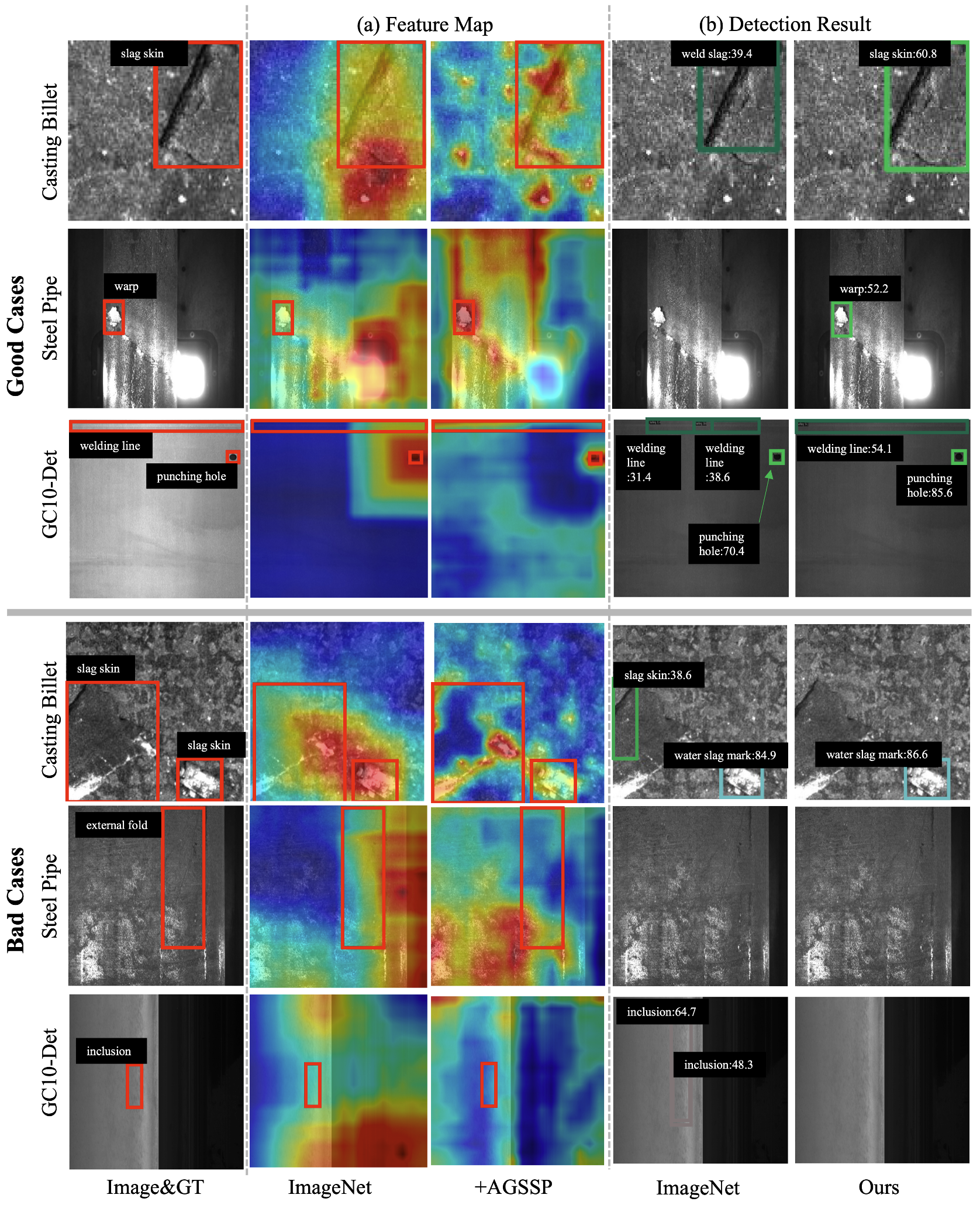}
\caption{Comparative analysis of ImageNet vs. Industrial AGSSP pretraining: Good and Bad Case Studies. (a) Visualization of the feature maps from the last backbone layer. (b) Comparison of detection results. GT - Ground Truth.}
\label{fig:vis1}
\end{figure}

\begin{figure}[!t]
\centering
\includegraphics[width=1.0 \linewidth]{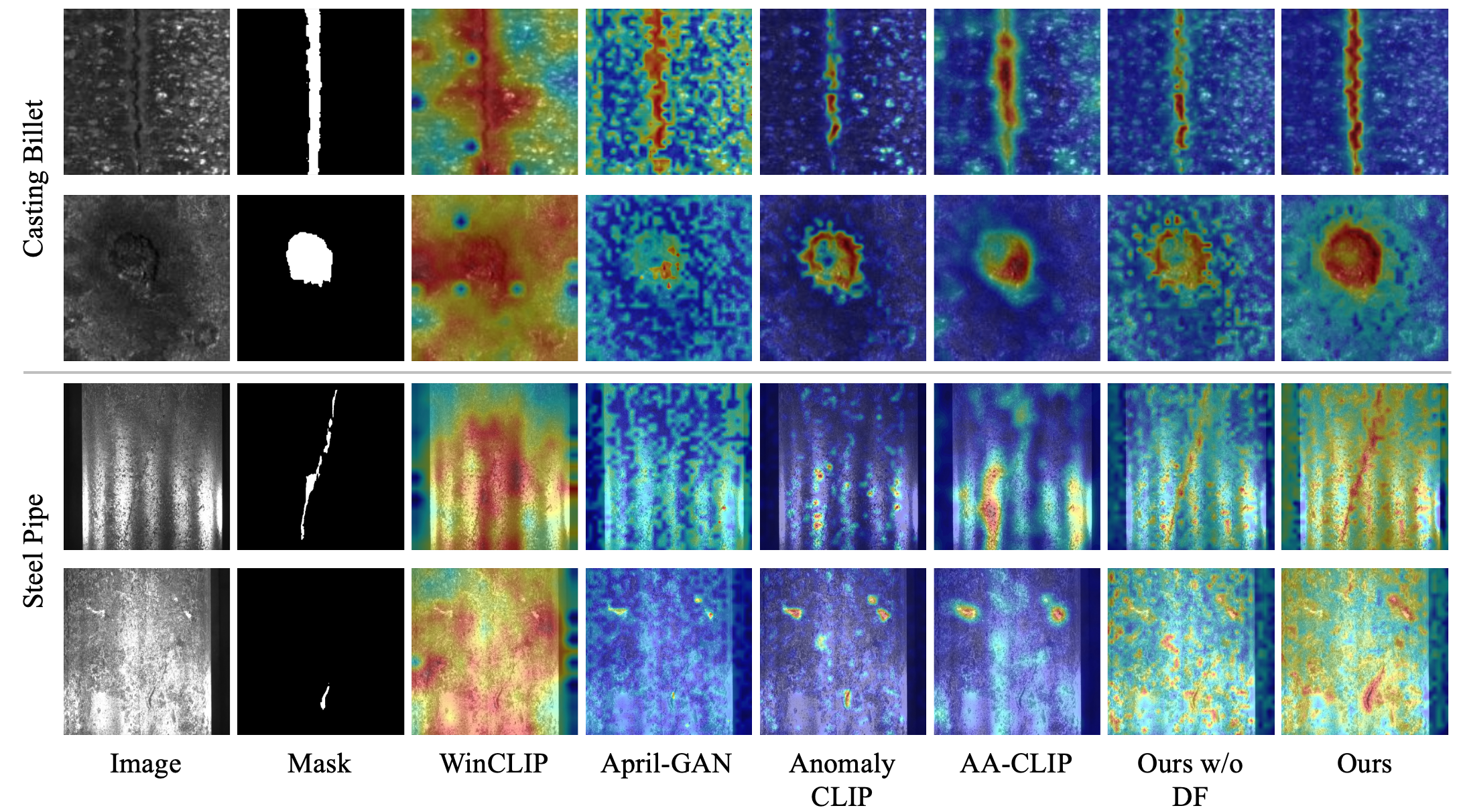}
\caption{Visual comparison of anomaly detection on complex industrial surfaces. \textcolor{markcolor}{Ours w/o DF denotes the generic-prompt ablation, replacing detailed defect descriptions with general anomaly prompts.} }
\label{fig:vis2}
\end{figure}

\subsubsection{Visualization Results}

Figure~\ref{fig:vis1} compares feature maps and detection results of the YOLOv8s-CSPDarknet model with and without anomaly-guided pretraining. The good cases show that pretraining enhances defect perception, yielding more precise localization and higher confidence scores. However, analysis on bad cases reveals inherent limitations in challenging scenarios where background noise mimics defect patterns or defects are extremely subtle. \textcolor{markcolor}{Since our anomaly priors are generated by few-shot vision-language models, they are inevitably imperfect. In particular, they may occasionally mistake normal reflections or complex textures for anomalies, or fail to highlight structural defects that lack strong semantic cues. Such noisy priors may misguide pretraining, causing the model to attend to pseudo-defective artifacts while overlooking faint but genuine defects, which leads to both false positives and missed detections. These observations highlight the need for more discriminative representation learning for industrial inspection under noisy anomaly priors. More specifically, anomaly-guided pretraining should not be expected to fully resolve the defect-versus-noise ambiguity under unlabeled supervision; rather, its role is to help the backbone learn finer-grained representations for both real defects and defect-like industrial noise. In this way, pretraining can better capture their subtle differences, while the subsequent supervised finetuning stage further determines which patterns correspond to actual defects and which should be suppressed as noise.} Despite these remaining challenges, the good cases still illustrate that our core strategy—using anomaly maps to guide the model’s attention—effectively improves the model’s perception of anomalous regions. This is also consistent with the overall quantitative gains achieved by our method. 

For anomaly detection, Figure~\ref{fig:vis2} illustrates that existing SOTA methods struggle with complex industrial surfaces and often mistake surface textures or lighting for anomalies. This highlights the need to incorporate additional defect knowledge. As shown in the comparison, using only generic prompts (Ours w/o DF) fails to distinguish subtle defects from background noise. In contrast, adding detailed defect descriptions enables our method to leverage the surrounding context, accurately localizing fine-grained defects. These visual results suggest that the knowledge-enhanced techniques facilitate precise detection of subtle defects even in noisy, complex scenarios.

\subsection{Ablation Studies}

Unless otherwise specified, all ablation studies were conducted on the Casting Billet dataset, using the Swin-Base model pretrained with SimMIM as the default baseline. 

\begin{figure}[!t]
\centering
\includegraphics[width=\columnwidth]{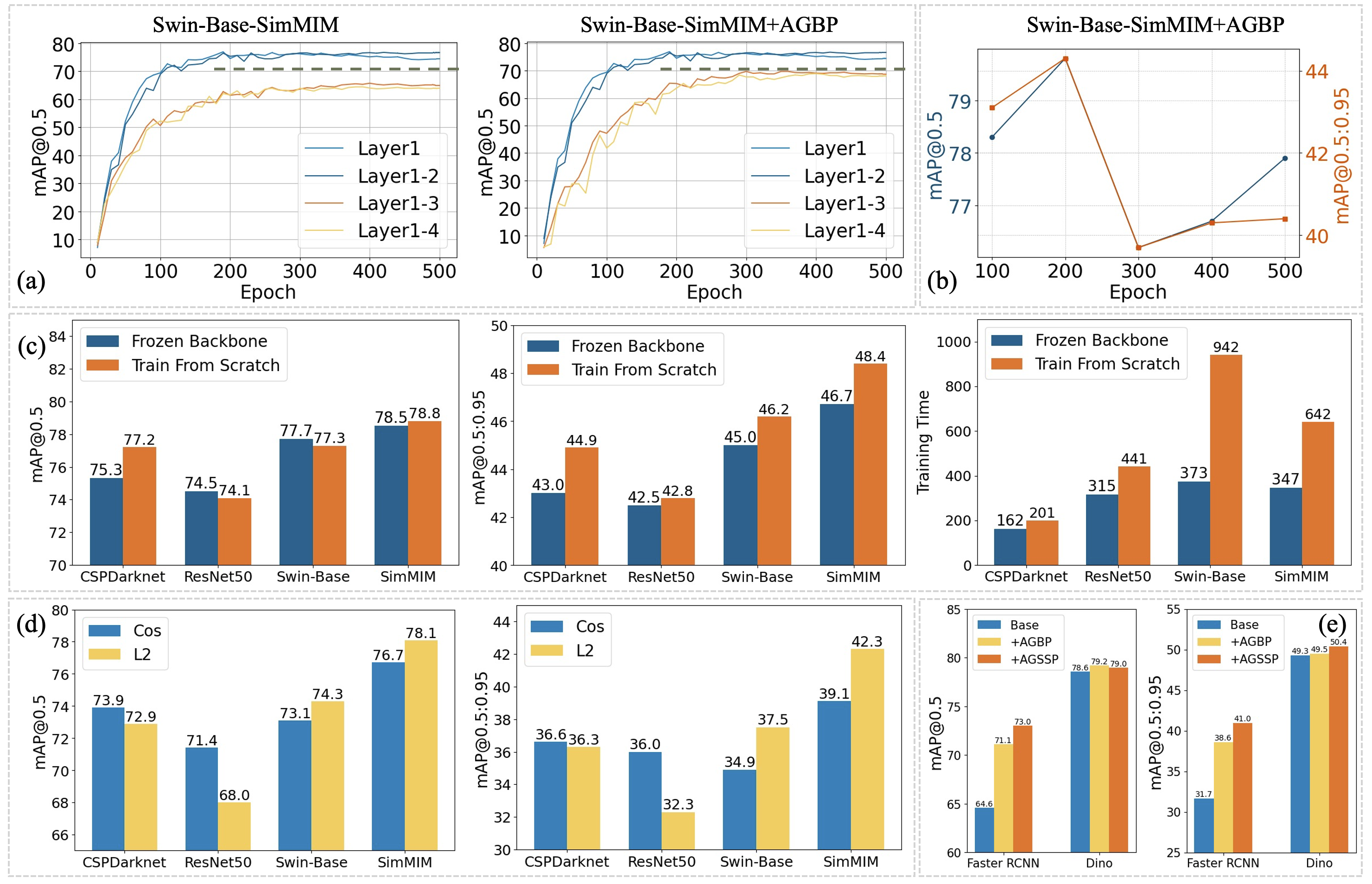}
\caption{(a) Effects of freezing layers 1–4 during finetuning. (b) Impact of different backbone pretraining epochs. (c) mAP and training time comparison: freezing vs. not freezing the backbone during detector pretraining. (d) Performance comparison: Cosine Distance vs. $L_2$ loss. (e) Ablation study on Faster RCNN \& Dino.}
\label{fig:exp}
\end{figure}

\subsubsection{Choice of Backbone Pretraining Hyperparameters}
This section explores the impact of backbone pretraining hyperparameters, specifically the number of epochs and the necessity of freezing lower layers. As depicted in Figure~\ref{fig:exp}(b), optimal results are achieved with 200 epochs of training. 
Initially, performance improves with increased training epochs, but after a certain point, it starts to decline.
One possible reason for this decline is overfitting to irrelevant patterns, which reduces the model's generalizability to downstream tasks. 
During backbone pretraining, we choose to freeze the lower layers (the first two layers) based on findings from the finetuning process. Specifically, as shown in Figure~\ref{fig:exp}(a), freezing the first two layers during finetuning yielded stable results, while freezing the third layer led to a noticeable performance drop. 
This indicates that lower-layer features tend to be universal and domain-independent, while higher-layer features are more task-specific and shaped by the pretraining objective. The performance degradation thus reflects a mismatch between the pretraining and finetuning tasks. By contrast, when AGBP is applied, freezing the final two layers produces significant performance gains compared with the case without AGBP, suggesting that integrating anomaly information enhances the model’s ability to learn transferable and task-relevant features for downstream defect detection. The results shown in Table~\ref{tab:ablation_study} further prove that freezing the lower layers during backbone pretraining enhances performance.

\begin{table}[]
\centering
\fontsize{8}{10}\selectfont
\setlength{\tabcolsep}{1.0mm}{

\begin{tabular}{c|c|c|cc}
\toprule
\multicolumn{1}{c|}{Backbone} & \multicolumn{1}{c|}{\begin{tabular}[c]{@{}c@{}}Frozen \\ Layers\end{tabular}} & \multicolumn{1}{c|}{\begin{tabular}[c]{@{}c@{}}Multi \\ Layers\end{tabular}} & \multicolumn{1}{c}{mAP@.5$\uparrow$} & \multicolumn{1}{l}{mAP@.5:95$\uparrow$} \\ \midrule
\multirow{4}{*}{\begin{tabular}[c]{@{}c@{}}Swin-Base\\ (SimMIM)\end{tabular}} 
& \XSolidBrush & \XSolidBrush  & 78.1 & 42.3 \\ \cline{2-5}
& \XSolidBrush & \Checkmark  & 78.1 & 43.4 \\ \cline{2-5}
& \Checkmark & \XSolidBrush  & 78.9 & 44.1 \\ \cline{2-5}
& \Checkmark & \Checkmark & \textbf{79.8} & \textbf{44.3} \\ 
\bottomrule
\end{tabular}
}
\caption{Ablation study on the freezing and distillation strategies during backbone pretraining. Frozen Layers refers to freezing the first two layers, while Multi Layers denotes distilling the anomaly map into the last two layers rather than only the final layer.}
\label{tab:ablation_study}
\end{table}

\subsubsection{Choice of Distillation Strategy}

We conduct ablation experiments on both the distillation loss functions and the choice of distillation layers. As shown in Figure~\ref{fig:exp}(d), we compare Cosine Distance (Cos) and $L_2$-based loss across different backbone architectures. The results are in agreement with empirical observations: $L_2$ performs better when transferring between ViT architectures, likely because the magnitude of feature representations remains comparable within the same architecture family. By contrast, Cos proves more effective for transfers across heterogeneous architectures, as it emphasizes directional consistency in the feature space while being insensitive to scale differences across models. Furthermore, Table~\ref{tab:ablation_study} indicates that distilling the anomaly map into the last two layers yields superior performance compared with distilling only into the final layer.

\begin{table}[t]
    \centering
    \fontsize{8}{10}\selectfont
    \setlength{\tabcolsep}{4pt} 
    \begin{tabular}{ll llll}
        \toprule
        \multicolumn{1}{c}{Pretraining} & \multicolumn{1}{c}{KeAD} & \multicolumn{2}{c}{Casting Billet} & \multicolumn{2}{c}{Steel Pipe} \\
        \cmidrule(lr){3-4} \cmidrule(lr){5-6}
         \multicolumn{1}{c}{Dataset}& \multicolumn{1}{c}{Encoder} & mAP@.5 & mAP@.5:95 & mAP@.5 & mAP@.5:95 \\
        \midrule
        ImageNet & - & 78.3 & \textbf{44.3} & 77.5 & 40.7 \\
        \midrule
        \multirow{3}{*}{Industrial} & - & 76.5 & 41.5 & 77.5 & 40.1 \\
         & Base & 78.1\textcolor{agreen}{(+1.6)} & 43.8\textcolor{agreen}{(+2.3)} & 79.3\textcolor{agreen}{(+1.8)} & 42.6\textcolor{agreen}{(+2.5)} \\
         & Huge & \textbf{79.8\textcolor{agreen}{(+3.3)}} & \textbf{44.3\textcolor{agreen}{(+2.8)}} & \textbf{81.4\textcolor{agreen}{(+3.9)}} & \textbf{43.7\textcolor{agreen}{(+3.6)}} \\
        \bottomrule
    \end{tabular}
    \caption{Ablation study on the impact of anomaly map quality (KeAD encoder capacity) on AGBP performance. Parentheses denote gains over the Industrial baseline, and - indicates the setting without AGBP.}
    \label{tab:ablation_dataset_encoder}
\end{table}

\subsubsection{Impact of Anomaly Map Quality}
To assess the impact of anomaly prior quality on feature learning, we conducted an ablation study on the AGBP stage using KeAD encoders with different capacities (ViT-Base vs. ViT-Huge). Table~\ref{tab:ablation_dataset_encoder} shows that performance gains are achieved when the quality of anomaly maps is improved. Specifically, the ViT-Huge encoder achieves a +2.1\% mAP@0.5 advantage over ViT-Base on the Steel Pipe dataset. 
Nevertheless, it is worth noting that even the less accurate ViT-Base encoder significantly outperforms the baseline. This observation confirms that while high-precision priors enhance discriminative learning, our framework exhibits strong robustness, effectively utilizing even noisy supervision signals to guide the model. 

\subsubsection{Freeze Backbone in Detector Pretraining}

This section compares the performance and total training time of freezing vs. non-freezing of the backbone during detector pretraining across various pretrained backbones. As depicted in Figure~\ref{fig:exp}(c), training from scratch yields a minor performance enhancement, up to 2\%. Nevertheless, freezing the backbone already offers notable performance benefits over ImageNet pretraining and it significantly shortens training time, presenting a practical solution depending on particular requirements.

\subsubsection{Effects on Other Detectors}

While the previous experiments were conducted within the YOLOv8 framework, we further evaluate the compatibility and effectiveness of our method on two additional detectors, Faster R-CNN~\cite{ren2016faster} and DINO~\cite{zhang2023dino}. The results, shown in Figure~\ref{fig:exp}(e), indicate that both +AGBP and +AGSSP improve the detection performance in two detectors. In particular, the improvement is more significant for Faster RCNN, which exhibits relatively lower baseline performance, boosting the mAP@0.5 from 64.6\% to 73\%, mAP@0.5:0.95 from 31.7\% to 41\%.

\begin{figure}[!t]
\centering
\includegraphics[width=\columnwidth]{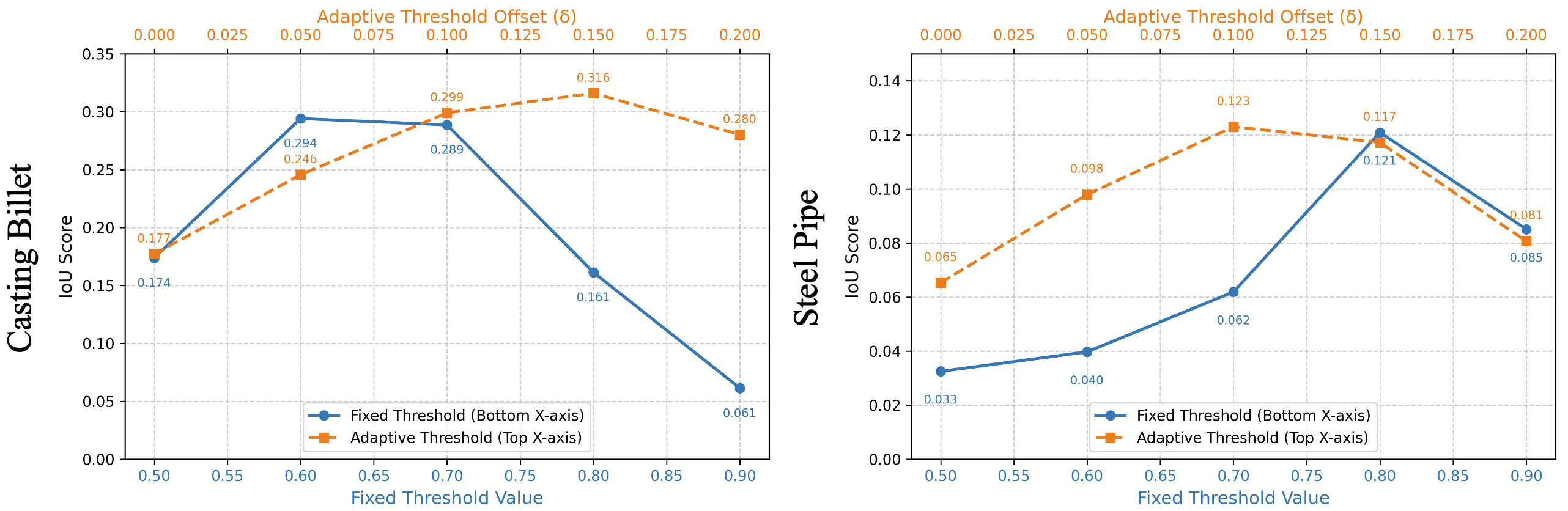}
\caption{An ablation study of $\delta$ parameter choices.}
\label{fig:exp_ano}
\end{figure}

\subsubsection{Choice of the Constant $\delta$}
Our ablation study on threshold selection (Figure~\ref{fig:exp_ano}) demonstrates that, compared to fixed thresholds require dataset-specific optimization, our adaptive method with $\delta=0.1$ achieves good performance across both datasets. It achieved higher mask IoU scores than even the optimally tuned fixed thresholds for each dataset.

\begin{table}[!t]
\centering
\fontsize{8}{10}\selectfont
\setlength{\tabcolsep}{1.0mm}{

\begin{tabular}{c|c|c|c|c|c}
\toprule
Model  & 
mAP@.5 & mAP@.5:95 & 
Params & GPU Load & FPS \\
\midrule
YOLOv8s  & 75.1 & 43.9 & 11.1M & 1851MiB  & 94.5 \\
YOLOv11s   & 74.7 & 45.2 & 36.0M & 531MiB & 111.4 \\
YOLOv12s   & 77.2 & 47.1 & 35.3M & 535MiB & 70.5 \\
RT-DETR   & 73.1 & 45.4 & 125.2M & 2797MiB & 26.5 \\
SLF-YOLO   & 72.6 & 38.6 & 10.2M &  368MiB & 78.6 \\
\midrule
CSPDarknet*  & 77.2 & 44.9 & 11.1M & 1851MiB  & 94.9 \\
Swin-Base* & \textbf{78.8} & \textbf{48.4} & 107.4M & 2843MiB & 23.4 \\
\bottomrule
\end{tabular}}
\caption{Comparative study with SOTA methods: Performance, Params (Parameters), GPU Load (GPU Memory Usage) and FPS(Batch Size = 1). * means our method.}
\label{tab:sota}
\end{table}

\subsubsection{Performance and Efficiency Comparison with SOTA Defect Detection Methods}

As quantitatively demonstrated in Table~\ref{tab:sota}, our method achieves SOTA trade-offs between accuracy and efficiency across multiple metrics, including mAP, model parameters, GPU memory usage, and FPS (averaged frames-per-second). Notably, the AGSSP-pretrained YOLOv8s model (CSPDarknet*) matches the detection performance of YOLOv12s while maintaining real-time capability (FPS $\geq$ 70). When implemented with Swin-Base backbone, our approach attains the highest mAP among all compared methods while achieving comparable inference speed to RT-DETR. It is important to note that our anomaly-guided pretraining method demonstrates two significant advantages. First, the generated pretrained weights exhibit strong transferability across diverse downstream applications while fully preserving the inference efficiency of deployed models. Second, while current validation focuses on YOLOv8, the framework is architecture-agnostic and can be easily extended to newer detectors like YOLOv12.

\section{Conclusion}
\label{sec:conclusion}

In this work, we collected a large-scale, unlabeled industrial dataset and evaluated the performance of several existing pretrained models. Contrary to expectations, we found that directly pretraining these models on industrial datasets led to performance degradation. Therefore, we introduced a novel self-supervised pretraining approach specifically designed for industrial defect detection tasks, called Anomaly-Guided Self-Supervised Pretraining (AGSSP). AGSSP leverages anomaly map priors to guide representation learning through a two-stage framework: anomaly map guided backbone pretraining and anomaly box guided detector pretraining. By using industrial defect descriptions to improve a few-shot anomaly detection method for generating anomaly maps, we identified potential defect information and distilled it into the pretraining process, encouraging the model to learn more defect-relevant feature representations. Experimental results show that AGSSP achieves state-of-the-art performance, robustly excelling across diverse datasets, backbones, and detector frameworks, and markedly outperforming models pretrained on ImageNet and COCO. In the future, we aim to study how dataset diversity and imbalance influence pretraining performance, and explore noise-invariant representation learning to better handle imperfect anomaly priors. 

\section{Acknowledgment}
This work was supported in part by the Beijing Science and Technology Planning Project, China (Z221100005822012), and Key Technologies Research and Development Program of China (2021YFB3202403). The computing work is supported by USTB MatCom of Beijing Advanced Innovation Center for Materials Genome Engineering.




\bibliographystyle{elsarticle-num}
\bibliography{ref_no_mark}
\end{document}